\documentclass[sigconf]{acmart}

\usepackage{booktabs} 

\setcopyright{rightsretained}
\usepackage{multirow}
\usepackage{subfigure}

\copyrightyear{2021} 
\acmYear{2021} 
\setcopyright{acmcopyright}\acmConference[SIGIR '21]{Proceedings of the 44th International ACM SIGIR Conference on Research and Development in Information Retrieval}{July 11--15, 2021}{Virtual Event, Canada}
\acmBooktitle{Proceedings of the 44th International ACM SIGIR Conference on Research and Development in Information Retrieval (SIGIR '21), July 11--15, 2021, Virtual Event, Canada}
\acmPrice{15.00}
\acmDOI{10.1145/3404835.3462839}
\acmISBN{978-1-4503-8037-9/21/07}

\begin{document}
\fancyhead{}

\title{Learning to Ask Appropriate Questions in Conversational Recommendation}

\author{Xuhui Ren}
\orcid{1234-5678-9012}
\affiliation{%
	\institution{The University of Queensland}
	\postcode{4072}
}
\email{xuhui.ren@uq.net.au}

\author{Hongzhi Yin*}
\thanks{*Corresponding author and having equal contribution with the first author}
\affiliation{%
	\institution{The University of Queensland}
	\postcode{4072}	
}
\email{h.yin1@uq.edu.au}

\author{Tong Chen}
\affiliation{%
	\institution{The University of Queensland}
	\postcode{4072}	}
\email{tong.chen@uq.edu.au}

\author{Hao Wang}
\affiliation{%
	\institution{Alibaba AI Lab}
}
\email{cashenry@126.com}

\author{Zi Huang}
\affiliation{%
	\institution{The University of Queensland}
	\postcode{4072}	}
\email{huang@itee.uq.edu.au}

\author{Kai Zheng}
\affiliation{%
	\institution{University of Electronic Science and Technology of China}
}
\email{zhengkai@uestc.edu.cn}

\begin{abstract}
Conversational recommender systems (CRSs) have revolutionized the conventional recommendation paradigm by embracing dialogue agents to dynamically capture the fine-grained user preference. In a typical conversational recommendation scenario, a CRS firstly generates questions to let the user clarify her/his demands and then makes suitable recommendations. Hence, the ability to generate suitable clarifying questions is the key to timely tracing users' dynamic preferences and achieving successful recommendations. However, existing CRSs fall short in asking high-quality questions because: (1) system-generated responses heavily depends on the performance of the dialogue policy agent, which has to be trained with huge conversation corpus to cover all circumstances; and (2) current CRSs cannot fully utilize the learned latent user profiles for generating appropriate and personalized responses.

To mitigate these issues, we propose the \textbf{K}nowledge-\textbf{B}ased \textbf{Q}uestion \textbf{G}eneration System (\textbf{KBQG}), a novel framework for conversational recommendation. Distinct from previous conversational recommender systems, KBQG models a user's preference in a finer granularity by identifying the most relevant \textbf{relations} from a structured knowledge graph (KG). Conditioned on the varied importance of different relations, the generated clarifying questions could perform better in impelling users to provide more details on their preferences. Finially, accurate recommendations can be generated in fewer conversational turns. Furthermore, the proposed KBQG outperforms all baselines in our experiments on two real-world datasets. 
\end{abstract}



\maketitle
\vspace{-0.2cm}
\section{Introduction}
With personalized recommendation services, e-commerce platforms can easily infer users' preference and generate personalized recommendations based on their interactions with the platform (e.g., searching, reviewing, and purchasing) \cite{gustavo}. Despite the great success achieved, traditional recommender systems are inevitably constrained by its requirement on the passively collected user feedback. This brings information asymmetry between users and recommender systems, as the system can only recommend items based on a user's history instead of her/his real-time intent whenever the service is used \cite{1145361913371769}.

Recently, the emerging language-based intelligent assistants such as Apple Siri and Google Home provide a new dimension in recommendation tasks. It enables an intelligent assistant to actively interact with users via conversations, so as to guide users to clarify their intent and find items that can meet their preferences \cite{33573843357939}. This possibility is envisioned as a holistic composition of a dialogue agent and a recommender system, and is termed as Conversational Recommender System \cite{1145361913371769,liu-etal-2020-towards}. In a general sense, CRS will proactively query a user with a series of clarifying questions to collect her/his demand on desired items. Then, the collected information serves as the preference context at the current timestamp, which helps generate timely and explainable recommendations. 

Increasing attempts to develop CRSs for e-commerce have been put forward in recent years, and most of them can be categorized into two types: (1) policy-based approaches \cite{10.1145/3209978.3210002, 96722939746, 3394592, 32692063271776, 33573843357939}; and (2) knowledge graph-based approaches \cite{wenqiangleikdd20, 33972713401180, NIPS2018_8180, chenetal2019towards}. Policy-based approaches originated from task-oriented dialogue systems, where a policy agent is designed to automatically react during the conversational recommendation (e.g., making a recommendation or querying the user's preference on one specific attribute type). Such methods can imitate the dialogue behaviors covered in the training corpus to generate human-like responses and complete the recommendation. However, their performance heavily relies on the labor- and knowledge-intensive process (i.e., acquisition and annotation) in preparing a comprehensive training dataset. Without enough well-annotated training corpus, most policy-based CRSs cannot learn an accurate dialogue action distribution for diversified dialogue state, thus underpeforming in dialogue generation. On the other hand, KG-based approaches have simplified the conversational recommendation into a ``System Ask, User Answer'' fashion with the help of knowledge graphs. By incorporating path-based reasoning into the conversation process, these methods can effectively offset the dependence on large training corpus. Each attribute node on the generated path will be used to formulate a yes/no question (e.g., ``Do you like pop/rock/country music?'') in order to affirm the user's current interest and explore the next node to connect. Finally, an explicit path in the KG will lead to a suitable item for recommendation. However, such exploration on KG only concerns the connectivity between nodes and overlooks the semantics behind users' responses. Apparently, they are less efficient as a node can be linked to multiple nodes, and the users may need to answer numerous questions before the correct item is reached.
\begin{figure}
	\setlength{\belowcaptionskip}{-0.5cm}
	\centering
	\includegraphics[width=0.9\linewidth]{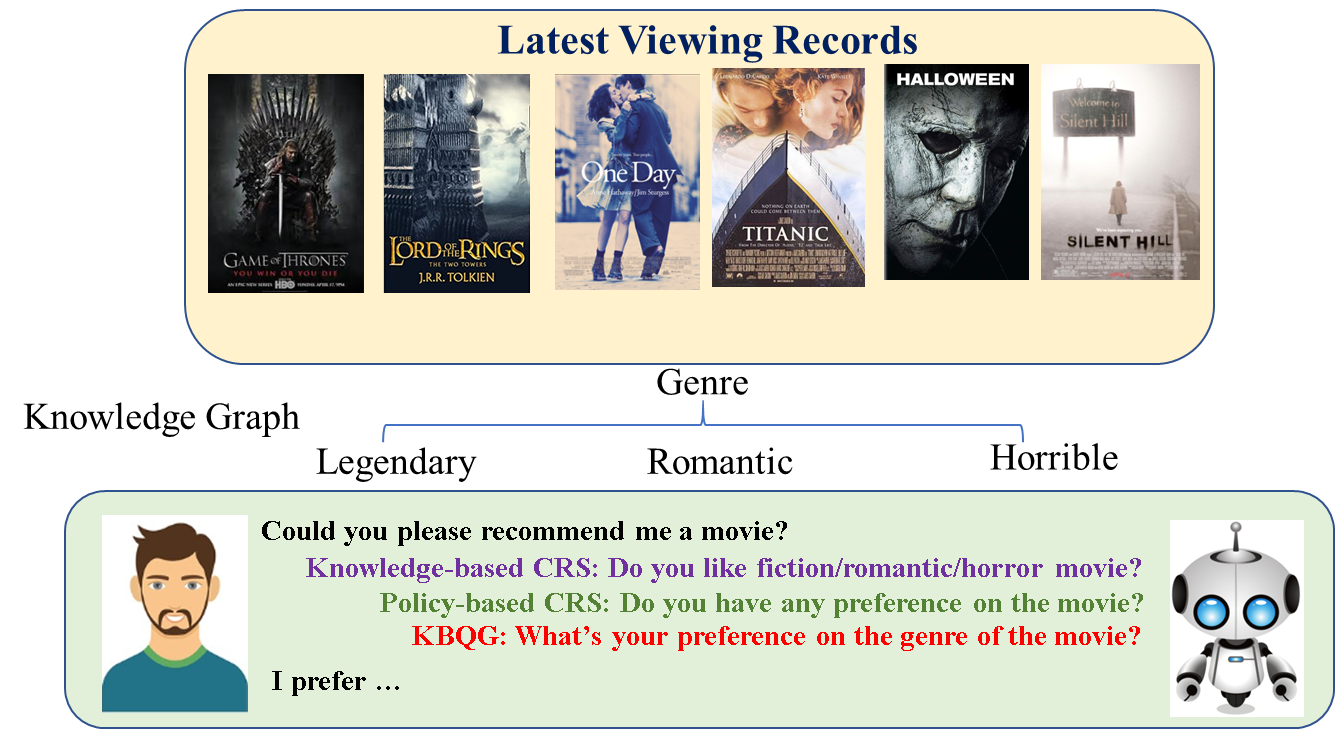}
	\caption{An example of clarifying questions for a conversational recommendation.}
	\label{fig:relation-type}
\end{figure}

Distinct from previous KG-based CRSs, we propose to fully account for the heterogeneous relations when reasoning over the paths in KG, so as to better understand the rationale behind a user's decision making process. As shown in Figure \ref{fig:relation-type}, the user recently has viewed several movies.  Specifically, we define attribute types (e.g., genre) as relations that connect items with different attribute values (e.g., romantic) With the help of the KG, the main genres of his watched movie can be categorized into three types: \textit{legendary}, \textit{romantic} and \textit{horror}. Our proposed KBQG could infer ``genre'' plays an important role in helping the user find the desired movie instead of other attribute types like ``actor'' and ``director''. Relations between interacted items are highly indicative on a user's intrinsic preferences, thus being beneficial for understanding fine-grained user interests. Therefore, our KBQG directly generates the clarifying questions to affirm the user's preference on the dominating attribute types for making an efficient and explainable recommendation. 
In light of this, we propose a novel framework named Knowledge-Based Question Generation system (KBQG) as a solution. Specifically, KBQG is designed to address the following key challenges in conversational recommendation tasks: 
\begin{enumerate}
\item \textit{What personalized question to ask?} Different users may focus on different aspects when consuming items. When generating clarifying questions, a CRS should prioritize critical item attribute types that influence the user's decision the most. This can make a CRS locate users' demand and avoid irrelevant questions with substantially fewer conversation turns. In KBQG, we devise a KG-based recommender component to infer each user's taste on different item attributes (represented via relations in KG) from historical interaction records. A more important relation means the associated attribute is more crucial to the user, which should be clarified at an early conversation stage to help the recommendation. 
\item \textit{Which system action should be performed?} Though the performance of the policy agent is directly associated with the recommendation quality, it is non-trivial to train an accurate policy agent given the potentially large action space. In this regard, we simplify our policy agent into a question-based paradigm, where the actions are either generating a recommendation or further asking a clarifying question. Before the recommendation module gains enough confidence, the CRS should continuously ask the user about her/his preference on different types of item attributes. Once the information is considered sufficient for an accurate recommendation, the CRS should execute the recommendation action.
\item \textit{How to react to users' responses?} The user will  give their responses after each turn of conversation. Hence, it is crucial to make full use of the contexts within users' responses to further advance the recommendation performance. In KBQG, we characterize two kinds of user responses. First, if a user provides an explicit affirmation on a system-generated clarifying question, KBQG can effectively identify the user's demand and store it for future recommendations. Second, once the user rejects the recommendation generated by the model, KBQG is able to generate more refined questions in order to further adjust its recommendation.
\end{enumerate}

To the best of our knowledge, KBQG's ability to generate personalized clarifying questions by investigating heterogeneous relations within KG is a brand new take on conversational recommendation. KBQG produces preference-aware clarifying questions, which help our system find users' desired items more efficiently. To validate the effectiveness of KBQG, we conduct extensive experiments on two real-world datasets, achieving state-of-the-art performance. Our contributions are summarized as follows:
\begin{itemize}
	\item We propose a novel framework KBQG for KG-based conversational recommendation, which models the user preference as well as the diverse KG relations for generating personalized, preference-aware clarification questions.
	\item We reformulate the conversational recommendation scheme and simplify the dialogue action space, releasing the heavy burden for learning a sophisticated policy agent.
	\item By simulating human conversations, we build two KG-based conversational recommendation datasets that can be a stepstone for future research. Experimental analysis fully showcases the superiority of the proposed KBQG.
\end{itemize}

\section{Related Work}
There are two lines of research inspired our proposed work: knowledge graph-based recommendation and conversational recommendation. We will discuss the research work in each relevant field.
\vspace{-0.3cm}

\subsection{Knowledge Graph-based Recommendation}
The great success of recommender systems makes them prevalent in e-commerce platforms to offer recommendations that can meet users' preferences. To predict the user preference, many approaches have been proposed based on collaborative filtering (CF), which assume behaviorally similar users exhibit similar appetites on items \cite{1033311843331267, 33972713401141, 8731605, 1032925003330877, 8839750,DBLP:conf/sigir/Lu0CNZ19, yin2016spatio}. However, CF-based methods usually perform poorly in modelling side information, such as user profiles and item attributes, thus cannot achieve satisfying performance when users and items have few interactions \cite{1011453292500330989}. In real-world applications, the various side information, including reviews, relational data and knowledge graphs (KGs) is important to a model's interpretability, and is also closely associated with the efficiency and persuasiveness of recommender systems \cite{1011453313705}. Incorporating side information into recommender systems to generate explainable recommendations has attracted much attention over the last few years, in which KG-based methods show significant performance advancements as KGs can provide rich facets about items.

Based on how graph-structured information is leveraged for recommendation, KG-based recommenders can be roughly categorized into three types, namely path-based \cite{101145321981932195, 101145240323240361, 10333113331203, 10635167311}, propagation-based \cite{10532925003330705,101142692063271739, 1011453292500330989} and embedding-based \cite{2886624, jietal2015knowledge, 1011453313705, 102939672939673, NIPS2013_1cecc7a7} methods. Path-based methods mainly explore various path connectivity patterns among entities in the KG to provide additional guidance for recommendation. The main effort for path-based methods is to deal with the vast amount of paths between entities, which is either solved by devising path selection algorithms \cite{101145240323240361, 1033311843331267} or by defining meta-path patterns \cite{101145321981932195}.
Meanwhile, propagation-based methods concentrate on information propagation over the whole KG to augment the information for recommendation. It strengthens entities' semantic representations via the embeddings of their high-order neighbours in KG \cite{101142692063271739, 10532925003330705}. The augmented entity embeddings serve as the exploitation of user interests or item properties, and are adopted for modelling user-item interactions \cite{1011453292500330989}. 
Besides, embedding-based methods mainly project entities and relations of a KG into low-dimensional continuous vector space, in which the rich structured knowledge is preserved to enhance the expressiveness of user and item representations. Despite the great success in KG-based recommenders, few of them can explain which factor influences a user's decision. \cite{1011453331184331188} models the user preference in a fine-grained manner and devises an attention-based neural network to analyze the influence of different relations on the user decision. However, similar to traditional recommenders, it assumes a user's interest is static over the long-time horizon \cite{wenqiangleikdd20}, which hinders the recommender from generating accurate recommendations to satisfy users' dynamic preferences.
\vspace{-0.2cm}

\subsection{Conversational Recommendation}
Conversational recommender systems are proved to be an effective solution for explainable and accurate recommendation to meet users' dynamic demand. 
This emerging research topic originates from task-oriented dialogue systems that help the user find her/his desired information with human-like dialogues \cite{10.1145/3209978.3210002}. Through multi-turn conversations with users, the dialogue system collects the user's intent and use a retrieval/content-based recommendation method to generate recommendations directly. The main challenge for this kind of method is to train an accurate policy agent to guide the dialogue action during the conversation.
Several approaches have been proposed in the last three years for training a policy agent for conversational recommendation \cite{10.1145/3209978.3210002, 96722939746, 3394592, 32692063271776, 33573843357939, 1033972713401042}. 
However, the training process for these approaches usually suffers from the scarcity of training data. Also, a successful conversational recommendation requires a policy agent to cope with diverse dialogue actions, which brings even more difficulties to the training process. 

Conditioned on the drawbacks of policy-based approaches, a new trend that employs KGs as an auxiliary information source for conversational recommendation has attracted much attention \cite{wenqiangleikdd20, 33972713401180, NIPS2018_8180, chenetal2019towards, adapting2021}. Early work on KG-based CRSs directly generates the recommendation based on connected neighbours of the user's mentioned items in the KG during the conversation \cite{chenetal2019towards, NIPS2018_8180}. 
They can provide users with recommendations and persuasive reasons in a single conversation turn. \cite{wenqiangleikdd20} applies the KG-based method into a multi-round conversation scenario. It utilizes the structural information of KG to query the user with yes/no questions about her/his explicit preference on the specific attribute value (i.e. entity). Through a series of questions and answers between the user and the dialogue agent, the CRS finds a recommendation path in KG for the final recommendation. Even though the affirmative answer from the user could help CRSs find a promising path in KG for recommendation, the entities in a KG may have multiple relations with other nodes, hence the user may need to answer numerous questions before reaching the correct recommendation. This defect of exisiting KG-based recommenders would significantly weaken and recommendation efficiency and the user experience.
\vspace{-0.3cm}
\section{Prelimiaries} 
Following \cite{wenqiangleikdd20}, we subsume our model under the multi-round conversational recommendation (MCR) paradigm but make some adjustments to improve its real-world applicability. MCR refers to one trial of recommendation as a round, and the CRS is free to interact with the user via attribute-related questions or recommendations multiple times until the task is finished. In this paper, we point out that asking generic questions can better impel the user to describe her/his specific demand compared with querying on explicit attributes. For example, asking ``What is your preference on the genre of the movie?'' is always more efficient and natural than blindly asking ``Do you like romantic/horrible/art/...?'' in capturing user preferences. So, we bypass the trial-and-error interaction style of the existing MCR paradigm, thus significantly shrinking the candidate question pool (i.e., action space) and making it more practical for e-commerce applications where a large amount of attributes are usually involved.

Specifically, an item is associated with a set of attribute types (e.g., color) and concrete attributes (e.g., blue) to describe its property. 
In a KG, an item is connected to an attribute value via an attribute type, e.g., $\textnormal{\textit{Jane Eyre}} \xrightarrow[]{genre} romantic$, where the attribute type (i.e., genre in this case) is regarded as a \textit{relation} between these two entities. An item may have multiple descriptions under a same attribute type, e.g. the genre of book \textit{Harry Potter} can be described by \textit{adventure}, \textit{fantasy} and \textit{fiction}. In a conversational recommendation round, KBQG aims to identify a user's most cared about attribute types (i.e., relations), then obtain her/his fine-grained preferences (i.e., attribute values) towards them by asking customized questions. The obtained preference information will eventually facilitate accurate recommendation. 

\begin{figure*}
	\includegraphics[width=0.95\linewidth]{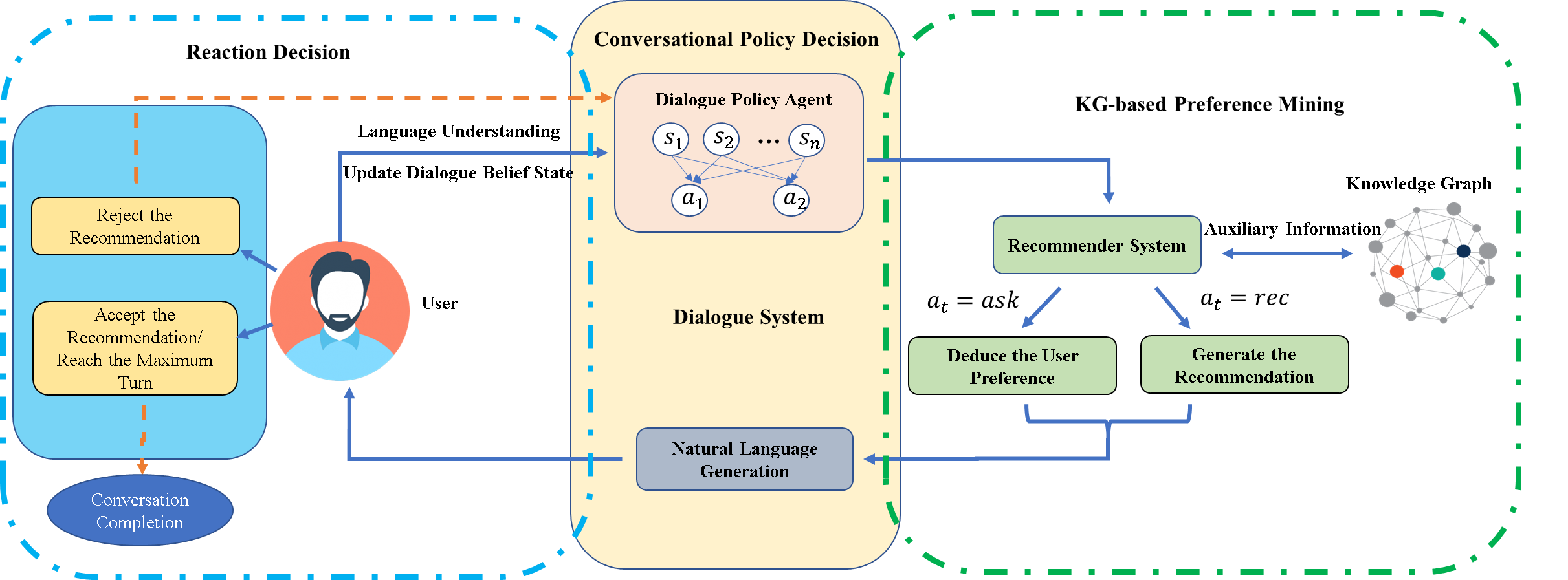}\Description{Several flies, spanning two
		columns of text}
	\caption{The overall diagram of KBQG. KBQG is executed with a three-phase process: Conversational Policy Decision, KG-based Preference Mining, and Reaction Decision.}
\label{fig:framework}
\end{figure*}

\section{Methodology}
As shown in Figure \ref{fig:framework}, our proposed KBQG is executed with a three-phase process consisting of conversational policy decision, preference mining, and reaction decision. At the beginning, the process is started by a recommendation request from a user. Conditioned on the current dialogue context, a policy agent is responsible for deciding the response action $a_t$ in the current conversation turn $t$, that is, whether to ask a clarifying question or to give a recommendation. The action $a_t$ will be passed onto the preference mining phase. If the policy agent decides to ask a clarifying question, the model will infer which relation in the KG is more important to the user from her/his interaction records, and then leverage the identified relations to guide the generation of the question. If the policy agent decides to provide a recommendation, then the preference mining module acts as a recommender to determine the items to be recommended based on the current dialogue belief state $\mathbf{b}_t$, which is a vector storing the user's current interests expressed during the conversation. 
After the user gives feedback to the system-generated response, KBQG will decide how it should react to the user. If the user provides explicit expressions for a clarifying question, KBQG will keep extracting the user demand and updating the dialogue belief state for conversational policy decision. Once the user accepts the recommendation, or the conversation reaches the maximum conversation turn, the conversation will be terminated; otherwise, if the user rejects the recommendation while the maximum conversation turn is not reached, our KBQG should return to the question-asking phase to generate further clarifying questions to amend the recommendation results. In what follows, we will present the design of each phase in detail.

\subsection{Conversational Policy Decision}\label{sec:CPD}
The policy decision component decides which dialogue action should be conducted in the current conversation turn. As discussed earlier, existing KG-based CRSs commonly face the difficulty in training a sophisticated policy agent to cover diversified dialogue situations. The main reason is that, each dialogue action is defined as one yes/no question about a specific attribute value (e.g., ``Do you like \underline{blue}?''), which leads to an unnecessarily large action space for decision making. It requires to train an estimation module to estimate the information entropy of each attribute value, deducing which attribute value should be selected as the next dialogue action. However, such an estimation module usually cannot obtain satisfying performance when the dataset contains a large quantity of sparse attributes, impeding both the prediction accuracy and model generalizability. Furthermore, it is not as efficient and natural as directly enquiring about the user at the attribute type level (e.g., ``Which \underline{color} do you like?''). 

We address this problem by simplifying the dialogue actions at the first place. Specifically, KBQG has only two actions, which are either asking clarifying questions or generating recommendations for the user. The questions asked by KBQG is fully tailored based on a users' attention towards different attribute categories, which is fulfilled by the subsequent preference mining component so that the policy agent can focus on choosing between two actions at each conversation turn. Thus, we design a policy agent $\pi(\cdot)$ with multilayer perception, which is formulated as: 
\begin{equation}
\pi_{\theta}(a_t|\textbf{s}_t) = \sigma(\textbf{W}_1 \cdot \tanh(\textbf{W}_2 \textbf{s}_t+\mathbf{b}_1) + \mathbf{b}_2),
\end{equation}
where $\sigma$ is the sigmoid activation function, and $\theta = \{\textbf{W}_1, \textbf{W}_2, \mathbf{b}_1, \mathbf{b}_2\}$ denotes the weights and biases to learn in the policy agent. $\textbf{s}_t$ is the internal dialogue state of the dialogue system that represents the current contexts of the dialogue at turn $t$, and it is defined as:
\begin{equation}
\textbf{s}_t = \textbf{b}_t \oplus \textbf{q}_t \oplus \textbf{c}_t ,
\end{equation}
where $\textbf{b}_t$ is the dialogue belief state which is a latent encoding of the user utterance (c.f. Section \ref{sec:reaction}), $\textbf{q}_t$ is a multi-hot vector indicating which attribute type has been clarified by the user, and $\textbf{c}_t$ is a vector indicating the ratio of candidate items whose affinity score with the user meets a predefined threshold $M$. The fewer candidate items means the recommender system has more confidence about the recommendation. The policy agent is trained in the RL for searching optimal conversational policy decisions. We adopt deep Q-learning as suggested in \cite{wenqiangleikdd20}, to allow for easier and faster convergence. Correspondingly, a reinforcement reward will be given after each conversation turn according to the user feedback. Intuitively, the policy agent needs to keep querying the user when: (1) few user preferences are captured (i.e., $\textbf{q}_t$ contains too many zeros); or (2) there are too many candidate items (i.e., $\textbf{c}_t$ represents a large ratio) which need to be narrowed down to ensure accurate recommendation. Naturally, the policy agent will perform the recommendation action when sufficient preference signals are collected and the candidate item pool is small yet dedicated to the information provided. Recall that we define each attribute type as a relation in our KG, and we design the reward in KBQG as follows: (1) $r_{t,p}$, if the user provides an explicit demand on the queried relation; (2) $r_{t,n}$, if the user currently does not provide any detail on the queried relation; (3) $r_{t,a}$, if the user accepts the recommendation; and (4) $r_{t,m}$, if the user refuses the recommendation or the conversation reaches the maximum turn with no successful recommendation. The RL agent selects one of the RL rewards as the evaluation reward $r_t$ to the current action. The policy agent is train towards maximizing the cumulated reward during the conversation. Finally, a policy gradient method is adopted to optimize the policy network via:
\begin{equation}
\bigtriangledown_{\theta}J(\theta) = \bigtriangledown_{\theta} \mathbb{E}[ \log\pi_{\theta}(a_t|\textbf{s}_t) R_{t}],
\end{equation}
where $R_t = \sum_{t'=t}^{T_{max}}\eta^{t'}r_{t}$ is the cumulated reward from turn $t$ to the end of conversation at $T_{max}$, and $\eta\in(0,1)$ is a damping factor that encourages the model to accomplish the task in fewer conversation turns.

\subsection{KG-based Preference Mining}\label{sec:preference_mining}
The preference mining module in KBQG has two main functionalities. First, when guiding the generation of clarifying questions, it infers the most important relations for the user within the KG, which are used for composing the questions. Second, when performing the recommendation, it is responsible for generating personalized recommendations based on the user preference. Different users tend to focus on different attribute types when shopping online, e.g., some are more concerned on the price while some have preference on specific brands. Such user preferences can be captured from users' historical interactions. To better account for each relation's impact on the user's decision process, we propose a translation-based approach in KBQG for preference mining.

Conventional translation-based methods model the user preference by learning an additional translation vector $\textbf{p}$ from a set of latent factors $\mathcal{P}$ for each user-item pair, 
such that given the representations of user $\textbf{u}$ and item $\textbf{i}$, we have $\mathbf{u}+\mathbf{p}\approx\mathbf{i}$ \cite{NIPS2013_1cecc7a7} if $u$ prefers $i$. The translation vector $\textbf{p}$ serves as a unique relation between users and items, and can be interpreted as the user's item-specific preference representation. The optimization objective is to minimize the $L1$ distance $||\mathbf{u}+\mathbf{p}-\mathbf{i}||$ for all positive user-item pairs.  
However, such method fails to consider the fine-grained user preference on attribute types when an interaction is made, hence are less suited for our goal of picking the most important attribute type for question generation.

Correspondingly, we design our preference mining module based on TransH \cite{105555289387394046}, which assumes each relation has its own hyperplane, and the preference-driven translation from the user to the item is valid only if they are projected onto the same hyperplane. Hence, in KBQG, we propose the hyperplane-based preference translation function as follows:
\begin{equation}\label{eq:trans}
f(u,i\mid p_u)=|| \mathbf{u}^{\perp}+\hat{\mathbf{p}}_u-\mathbf{i}^{\perp}||,
\end{equation}
where $\mathbf{u}^{\perp}$ and $\mathbf{i}^{\perp}$ are projected user and item representations, and $\hat{\textbf{p}}_u$ is the user's preference-driven translation vector. First, $\mathbf{u}^{\perp}$ and $\mathbf{i}^{\perp}$ are obtained via:
\begin{equation}
\begin{split}
\hat{\mathbf{u}} &= \mathbf{u}+\mathbf{b},\\
\mathbf{u}^{\perp} &= \hat{\mathbf{u}} - \mathbf{w}_{u}^{\mathrm{T}}\hat{\mathbf{u}}\mathbf{w}_{u},\\
\mathbf{i}^{\perp} &= \mathbf{i} - \mathbf{w}_{u}^{\mathrm{T}}\mathbf{u}\mathbf{w}_{u},
\end{split}
\end{equation}
where $\hat{\mathbf{u}}$ is the user embedding enhanced by the dialogue belief state $\mathbf{b}$, thus encoding the user's current preference during the conversation. Noticeably, to make this preference translation more suited to conversational recommendation, we further incorporate conversational contexts by augmenting the user embedding with dialogue belief state. $\mathbf{w}_{u}$ is the projection weight. Specifically, they are learned from the user's previously interacted items, where relations (i.e., attribute types) are thoroughly modelled to extend their semantic expressiveness. This also provides an explicit explanation for user's potential interactions with items. First, $\hat{\mathbf{p}}_u$ can be computed via:
\begin{equation}
\hat{\mathbf{p}}_u= \sum_{r\in \mathcal{R}}\alpha_{r,u} \mathbf{r},
\end{equation}
where $\mathbf{r}$ is the representation of each relation in the relation set $\mathcal{R}$, and $\alpha_{r,u}$ is user $u$'s attention weight on each specific relation. We define $\alpha_{r,u}$ with the standard softmax function:
\begin{equation} 
\alpha_{r,u}=\frac{\exp(a(r,u))}{\sum_{r'\in R} \exp(a(r',u))},
\end{equation}
where $a(r,u)$ is the attention score of user $u$ towards relation $r$. We learn it with an attention network:
\begin{equation} 
a(r,u) = \mathbf{h}^{\mathrm{T}}(ReLU(\mathbf{W}_3(\mathbf{u}\oplus\mathbf{r})+\mathbf{b}_3))
\end{equation}
$\mathbf{W}_3$ and $\mathbf{b}_3$ are weight matrix and bias vector, and $\mathbf{h}$ projects the calculated latent representation into a scalar score. A higher attention score means user $u$ exhibits higher interest on $r$ when searching for the next item to interact with. All relations are ranked in a descending order with their corresponding attention weights. KBQG sequentially pick an unused attribute type in each conversation turn for generating clarifying questions. As such, instead of learning a latent preference to connect items and users as in \cite{1011453313705}, we incorporate the rich semantics from the relations in KG to infer the user's most concerned attribute types, making it possible to generate appropriate questions for conversational recommendation. 

Similarly, the projection vector $\mathbf{w}_u$ can be obtained with the projection vector $\mathbf{w}_r$ on each relation hyperplane:
\begin{equation}\label{eq:rela_proj}
\mathbf{w}_{u}= \sum_{r\in R}\beta_{r,u} \mathbf{w}_{r},
\end{equation}
where another set of attention weights $\beta_{r,u}$ are computed, and  $\textbf{w}_r$ denotes the relation-specific projection weight. By replacing $\textbf{r}$ in Eq.6 with $\textbf{w}_r$ in a parallel attention network, we can calculate $\beta_{r,u}$ and $\mathbf{w}_{u}$ analogously. 

Thus, the auxiliary information from KG and dialogue content could enhance the semantic representation and learn the connections between users and items. For each user $u$, we define Bayesian Personalized Ranking \cite{journalc2618} loss to encourage $u$'s translation distance between all interacted items $i\in \mathcal{I}_u^{+}$ to be smaller than randomly selected negative items $i' \in \mathcal{I}_u^{-}$:
\begin{equation}
\mathcal{L}_f = \sum_{(u,i)\in \mathcal{H}^{+}} \sum_{(u,i^{'})\in \mathcal{H}^{-}} -\mathrm{ln}\big{(}\sigma(f(u,i'\mid p_u) - f(u,i\mid p_u))\big{)},
\end{equation}
where $\mathcal{H}^{+}=\{\mathcal{I}_u^+\}_{\forall u}$ and $\mathcal{H}^{-}=\{\mathcal{I}_u^-\}_{\forall u}$ are all positive and negative interaction instances used for training. In order to effectively preserve the relations between entities within the KG, we jointly optimize the recommendation loss with a relation modelling loss in KG. According to the assumption of TransH, the translation distance between the head entity $e_h$ and the tail entity $e_t$ is defined as:
\begin{equation}
f(e_h,e_t, r)=||\mathbf{e}^{\perp}_h+\mathbf{r}-\mathbf{e}^{\perp}_t||
\end{equation}
where the projected embeddings of the head and tail entities can be obtained via:
\begin{equation}
\begin{split}
\mathbf{e}^{\perp}_h &= \mathbf{e}_h - \mathbf{w}_r^{\mathrm{T}}\mathbf{e}_h\mathbf{w}_r\\
\mathbf{e}^{\perp}_t &= \mathbf{e}_t - \mathbf{w}_r^{\mathrm{T}}\mathbf{e}_t\mathbf{w}_r
\end{split}
\end{equation}
where $\mathbf{w}_r$ is the relation-specific projection vector as in Eq.\ref{eq:rela_proj}. The training objective of TransH is to maximize the margin between the observed triples in $\mathcal{KG}^+$ and randomly sampled false triples in $\mathcal{KG}^{-}$, which is defined as the following:
\begin{equation}
\mathcal{L}_g = \sum_{(e_h,e_t,r)\in \mathcal{KG}} \sum_{(e_h', e_t', r')\in \mathcal{KG}^{-}} [f(e_h,e_t, r) - f(e_h',e_t', r')].
\end{equation}

The final training objective function can be written as:
\begin{equation}
\mathcal{L} = \lambda \mathcal{L}_f+ (1-\lambda)\mathcal{L}_g 
\end{equation}
where $\lambda$ is used to trade-off the influence of two parts.

\subsection{Reaction Decision}\label{sec:reaction}
After the user gives corresponding feedback on the system-generated action (i.e. clarifying question or recommendation), KBQG is supposed to react accordingly based on the information carried by the user feedback. When KBQG has not collected enough information about the user interest, the user will respond to the generated questions to clarify her/his demand on specific attribute types. In this condition, we introduce an entity linking algorithm proposed by \cite{10114518714371871689} to accurately identify the descriptive phrases that appeared in the users' replies. These identified entities (i.e., detailed attribute values) $z_1, z_2, \dots, z_t$ are recognized as the user's current explicit interest on the attribute types asked, whose embeddings are used for composing the dialogue belief state $\mathbf{b}_t$: 
\begin{equation}
\mathbf{b}_t=\sum_{t'=1}^{t}\mathbf{z}_{t'},	
\end{equation}

When the policy agent decides to make recommendation, KBQG will search in the database with recommendation score $f(u,i \mid p_u)$ and output top-$K$ items (i.e., $K$ items with lowest $f(u,i,p_u)$ scores) for recommendation. The user can respond to the recommendation list with acceptance or rejection. If the user accepts the recommendation, this conversation is considered as a success; otherwise, our system needs to further collect user interest and narrow down the recommendation candidates until the maximum of the conversation turn $T$. Meanwhile, the failed recommendations are stored in a negative set $\mathcal{S}_u^{-}$ and are avoided in the next round of recommendation. That is, the item candidate set for $u$ is $\mathcal{S}_u^{cand} = \mathcal{S}_u^{+} \backslash \mathcal{S}_u^{-}$, where $\mathcal{S}_u^{+}$ is the set of candidate items whose predicted distance to $u$ is smaller than the threshold $M$ described in Section \ref{sec:CPD}. Finally, if the system does not find the user's desired item until the maximum conversation turn, we regard it as a failed recommendation attempt.

\subsection{Training}
The whole training process includes two parts: (1) offline training for preference mining; and (2) online training for conversational policy decision. The training objective for the first part is to minimize the distance between the embeddings of a user and her/his preferred items. A lower score of $f(u,i\mid p)$ indicates higher interest of the user to the item. For each user, we regard all entities connected to her/his interacted items as the user's affirmed preference, and store them in the $b_t$ to train the recommender. A multi-task training method is adopted to optimize both the recommendation and KG modelling losses as mentioned in Section \ref{sec:preference_mining}.

For online training, we introduce a user simulator to interact with KBQG to train the policy agent. The user simulator is a simplified version of the one used in \cite{wenqiangleikdd20}. In our case, the user simulator provides the user-side information based on the user's historical interactions. For each conversation session, the user's current preference is predefined by the attribute types and values associated with her/his interacted items in KG. The conversation between KBQG and the user simulator includes two parts. Firstly, KBQG sequentially asks clarifying questions during the conversation. If the queried relations match the user's preference on attribute types, the user simulator will respond with detailed attribute values. Secondly, when KBQG provides the recommendation list to the user, the user simulator will accept the results if the ground-truth item $i$ is present. Otherwise, the recommendation is rejected. 

\section{Experiments}
We evaluate KBQG on two real-world datasets. Specifically, we aim to answer the following research questions (RQs):

\noindent\textbf{RQ1:} How does our proposed KBQG compare with state-of-the-art conversational recommendation methods?

\noindent\textbf{RQ2:} Are our user preference inference scheme and simplified policy agent really effective?

\noindent\textbf{RQ3:} Can KBQG help comprehend user preferences and provide convincing yet explainable recommendation with conversations?

\subsection{Datasets}
We conduct experiments on two publicly available datasets from two different domains, namely MovieLens-1M for movie recommendation\footnote{https://grouplens.org/datasets/movielens/} and DBbook2014 for book recommendation\footnote{http://2014.eswc-conferences.org/important-dates/call-RecSys.html}. For both datasets, we follow \cite{1011453313705} to transform users' explicit ratings into implicit feedback where ratings meeting a predefined threshold are considered as positive interactions. The rating threshold is 4 for MovieLens-1M and is 1 for DBbook2014 due to data sparsity. For each user, we also sample an equal amount of negative items to match the positive interactions. 

To build KGs for both datasets, we map items to DBPedia entities (if available), and filter out relations that are infeasible for generating clarifying questions, e.g., non-textual data like URLs. 
We preprocess both datasets by filtering out infrequent users and items (i.e. retaining users and items with at least 10 interactions). Table 1 shows the statistics of two datasets after processing. There are 6,040 users, 3,240 items and 998,539 interaction records in MovieLens-1M dataset. On average, each user has 165 interaction records. DBbook2014 has 5,576 users, 2,680 items and 65,961 interaction records, leading to an average of 12 interactions per user. The KG triples used in both datasets are at the same scale, where the subgraph for MovieLens-1M composes of 159,492 triples with 87,492 entities and 13 relations, while the subgraph for DBbook has 194,710 triples with 8,793 entities and 9 relations. 
\begin{table}[]
	\caption{Statistics of MovieLens-1M and DBbook2014}
	\begin{tabular}{c|c|c|c}
		\hline
		\multicolumn{2}{c|}{}                                & MovieLens & DBbook2014 \\ \hline
		\multirow{4}{*}{Interactions} & Users       & 6,040         & 5,576       \\ \cline{2-4} 
		& \#Items       & 3,240         & 2,680       \\ \cline{2-4} 
		& \#Ratings     & 998,539       & 65,961      \\ \cline{2-4} 
		& \#Avg. Interactions
		& 165          & 12         \\ \hline
		\multirow{3}{*}{KG}        & \#Entity      & 9,457        & 8,793       \\ \cline{2-4} 
		& \#Relation    & 13           & 9          \\ \cline{2-4} 
		& \#Triple      & 159,492       & 194,710     \\ \hline
	\end{tabular}
\end{table}
\subsection{Experimental Settings}
\subsubsection{Training Details}
We randomly split each dataset for training, validation and test with the ratio of $7:2:1$. When making recommendations, we set $K=10$ for generating the top-$K$ item list. For each dataset, the maximum conversation turn is set to fit the total number of KG relations plus one for recommendation, i.e., $T=14$ for MovieLens-1M and $T=10$ for DBbook2014, respectively. The rationale is that, if a recommender has already obtained users' preferences on all relations, it should start making recommendations as no further information can be acquired from the user. 
We respectively set the hyperparameters for offline training and online training phases. For offline training, the embedding size is 100, and the batch size is 256. The optimizers for the recommendation module and KG modelling are Adagrad and Adam with the learning rate of learning rate of 0.003 and 0.001, respectively. For recommendation, we also impose an $L_2$ regularization weighted by $10^{-4}$. We set the trade-off coefficient $\lambda$ to 0.5 for MovieLens-1M and 0.7 for DBbook2014 following \cite{1011453313705}.  
For online training, the policy agent is trained with deep Q-learning using RMSprop. The specific values of RL rewards are designed as: $\{r_{t,p} = 0.1, r_{t,n}= -0.1, r_{t,a} = 1, r_{t,m} = -0.3\}$. The deep Q-learning has the batch size of 128, the experience memory size of 100,000, and the damping factor $\eta$ of 0.9. After performing grid search in $\{0.25, 0.5, 0.75\}$, we set the recommendation threshold $M$ in the dialogue belief state to 0.25 and 0.5 on MovieLens-1M and DBbook2014, respectively.

Considering generating extremely human-like responses is not the main research task in this paper, we currently design a series of response templates for constructing different clarifying questions. The selected relations are filled in the vacant position of the templates to conduct the conversation. We leave the exploration of language generation in the future work. \footnote{The source code will be release at https://github.com/XuhuiRen/KBQG.}

\subsubsection{Baselines}
Studies on CRS are emerging in recent years, which focus on different application scenarios. To verify the performance of our proposed KBQG, we select the state-of-the-art methods that have similar task settings as ours. The baselines are summarized as follows: 
\begin{description}
	\item[CRM:]
	This is a policy-based CRS that integrates a task-oriented dialogue system with a recommender  \cite{10.1145/3209978.3210002}. The main idea is to involve a semi-structured user query to represent the dialogue history as well as the user's current preference. This query is then used for generating personalized recommendations. 
	
	\item[EAR:]
	This method is designed in MCR scenario and is equipped with an Estimation-Action-Reflection framework for CRS \cite{1145361913371769}. It estimates the importance of each attribute with information entropy and utilizes a policy agent to select a suitable attribute from in each conversation turn to formulate a yes/no question, acquiring the user preference information for recommendation.
	
	\item[SCPR:]
	This method proposes to incorporate path searching on KG into CRS in MCR scenario \cite{wenqiangleikdd20}. It formulates a series of yes/no questions to collect the user's feedback and use these feedback as the guidance search on KG to find the recommendation.  
\end{description}
Although there are other recent CRSs \cite{chenetal2019towards, NIPS2018_8180, liu-etal-2020-towards,10339448634, 3394592}, they are inapplicable for comparison due to different task settings. For example, \cite{NIPS2018_8180, 10339448634} is designed to recommend a similar recommendation to a specific prototype without clarifying the user's real-time preference, and \cite{liu-etal-2020-towards, 3394592} are more concerned on language understanding and generation.

\subsubsection{Evaluation Metrics}
We follow \cite{wenqiangleikdd20} to evaluate the performance of each CRS with success rate at turn $T$ (SR$@T$) and average turn (AT) of conversations. SR$@T$ is the accumulative task success rate by turn $T$, and we set $T=T_{max}$ by default for overall performance comparison. AT is the average number of conversational turns needed for a successful recommendation session. If a user cannot find his desired item till $T_{max}$, we return $T_{max}$ for computing SR. A higher score on SR represents better recommendation accuracy, whereas a lower AT score indicates a model could more efficiently deliver a successful conversational recommendation using fewer conversations. 

\subsection{Performance Comparison (RQ1)}
Table 2 shows the overall performance of our proposed system as well as the baseline systems. It is obvious that our proposed system achieves the best performance among all methods on both datasets regarding SR and AT.
\begin{table}[]
	\caption{Performance comparison of all methods on two datasets by SR and AT, where the best performance is boldfaced ($p<0.01$ in one-sample paired $t$-tests when comparing all our results with baselines' scores).}
	\begin{tabular}{c|c|c|c|c}
		\hline
		\multirow{2}{*}{} & \multicolumn{2}{c}{DBbook2014} & \multicolumn{2}{c}{MovieLens-1M} \\ \cline{2-5} 
		& SR           & AT             & SR            & AT              \\ \hline
		CRM               & 0.286          & 8.593          & 0.761           & 8.117           \\ \hline
		EAR               & 0.247          &   9.369        & 0.594           & 10.362           \\ \hline
		SCPR              & 0.275          & 9.255          & 0.659           & 10.099           \\ \hline
		KBQG              & $\mathbf{0.323}$   & $\mathbf{8.156}$  & $\mathbf{0.816}$ & $\mathbf{7.335}$              \\ \hline
	\end{tabular}
\vspace{-0.5cm}
\end{table}

Compared with KG-based methods EAR and SCPR, CRM and our proposed KBQG exhibit significant advantages on both DBbook2014 and MovieLens-1M. The main reason behind this phenomenon is the items on both KGs for two datasets are linked with a lot of attributes. However, the expansion of the item attributes requires a huge amount of training set to train an estimation module that can estimate the information entropy on the whole attribute when generating the clarifying questions, which is a key component in EAR and SCPR. Without an accurate entropy estimation on the potential attributes, their system cannot find the suitable attributes to ask for the user preference, leading to a terrible performance on task completion and conversion efficiency. What's more, the clarifying mode for EAR and SCPR collects the user preference with a series of yes/no questions on the specific attributes. It inherently cannot achieve a satisfying efficiency when there is a large attribute pool to describe the items. Our proposed KBQG and CRM clarify the user preference with a set of generic questions. The user can describe his preference on each attribute type with the guidance of clarifying questions, achieving a better conversational recommendation experience.

Compared with policy-based method CRM, our proposed KBQG still presents remarkable improvement. The reason is that the policy agent of CRM mainly relies on the dialogue action distribution learned from the training corpus to generate the next clarifying question. It lacks the ability to analyze the importance weight of different relations on the user's decision. Therefore, it usually requires more user preference and more conversation turn for the recommendation; and the ultimate process of CRM is decomposed to a traditional recommender system Factorization Machine \cite{10.1145/3209978.3210002} to recommend the items. Hence, it fails to utilize the connection information on KG to enhance the entity embedding.

\begin{figure}[htbp]
	\setlength{\abovecaptionskip}{-0.1cm}
	\setlength{\belowcaptionskip}{-0.5cm}
	\centering
	\subfigure{}{
		\centering
		\includegraphics[width=3.5cm]{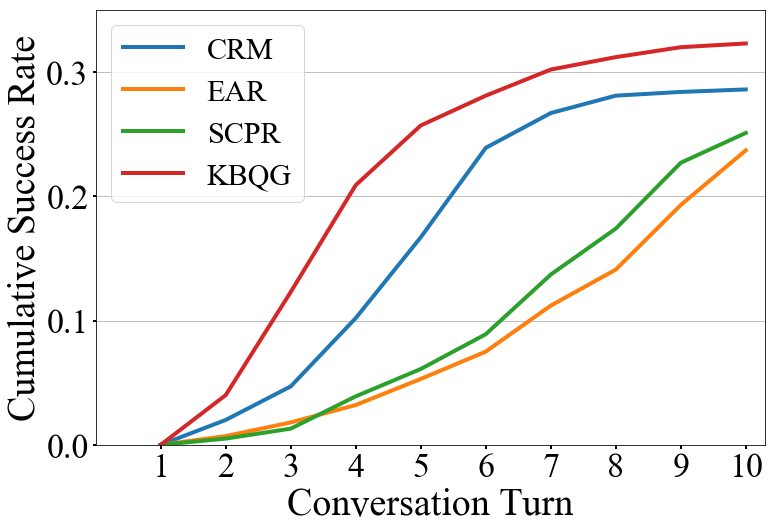}
	}
	\subfigure{}{
		\centering
		\includegraphics[width=4.5cm]{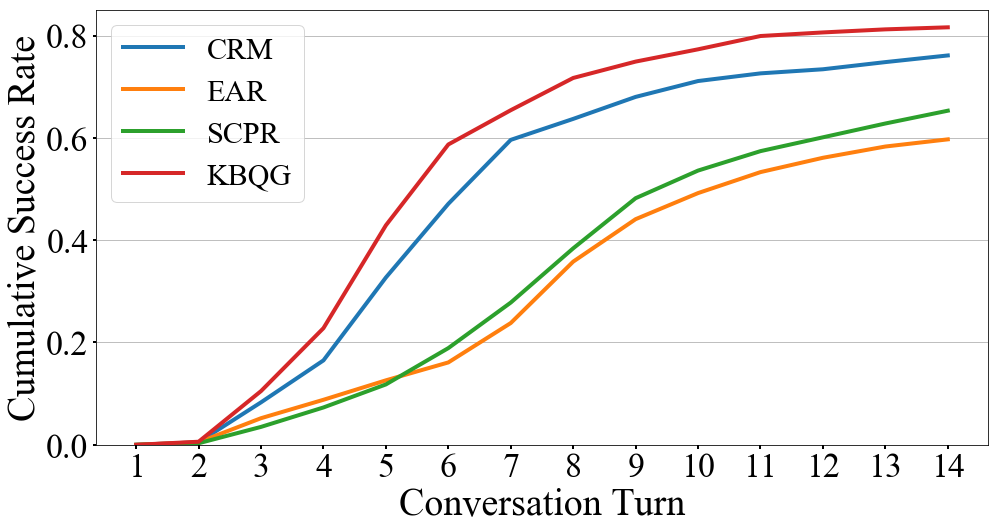}
	}
	\caption{Cumulative success rate of compared methods at different conversation turns on DBbook2014 (left) and MovieLens-1M (right).}
	\label{fig:full_result}
\end{figure}

\begin{figure*}
	\includegraphics[width=0.95\linewidth]{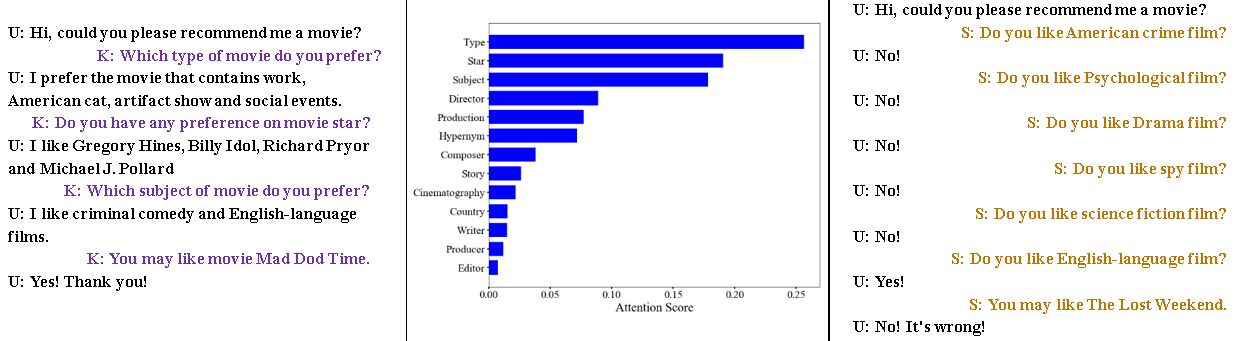}\Description{Several flies, spanning two
		columns of text}
	\caption{Sample conversations generated by KBQG (left) and SCPR (right) and attention scores of different relations (middle).}
	\label{fig:dialoguedemo}
\end{figure*}

Figure \ref{fig:full_result} presents a finer-grained performance comparison of the task cumulative success rate on two datasets. From the figure, we can observe that our proposed KBQG could reach a higher task success rate at the early stage of the conversation. The reason is that our proposed KBQG could inference which relation is more important for the user to make a decision, and give priority to the user preference on the relations that occupy a high attention weight for generating the recommendation. It can filter many unimportant relations during the clarifying period to accelerate the user preference collecting process. Also, the recommendation performance in our system is benefit from the relational modelling with TransH, and achieves better performance compared with a traditional recommender system.

\subsection{Analysis on Key Designs (RQ2)}
The key design in our proposed KBQG is the simplification and optimization of the conventional policy-based CRS. The policy agent in our proposed KBQG only accounts for deducing a dialogue action, asking a clarifying question or presenting the recommendation, which means the dialogue action space is greatly decreased. Hence, it can achieve superior performance on the dialogue action generation. KBQG could model the user decision policy from the rich resource of historical interactions, providing an effective way to infer which relation is more appropriate for formulating the next clarifying question. To verify its effectiveness, we conduct ablation experiments by designing two variants of our KBQG, marked with $\textbf{KBQG-P}$ and $\textbf{KBQG-A}$. KBQG-P replaces the current policy agent with a standard policy agent as in \cite{10.1145/3209978.3210002} that has $R+1$ dialogue actions. It decides the next dialogue action based the distribution of dialogue actions in the training corpus. The policy agent behaves similarly compared with other policy-based methods in each conversation turn when selecting the next relation and deciding asking/recommending. All other components in KBQG-P are inherited from KBQG. KBQG-A maintains the policy agent in KBQG, but replaces the attentive relation aggregation with average pooling, hence all relations share the same possibility of being selected to formulate the clarifying question. 
\begin{figure}[htbp]
	\setlength{\belowcaptionskip}{-0.5cm}
	\setlength{\abovecaptionskip}{-0.1cm}
	\centering
	\subfigure{}{
		\centering
		\includegraphics[width=3.5cm]{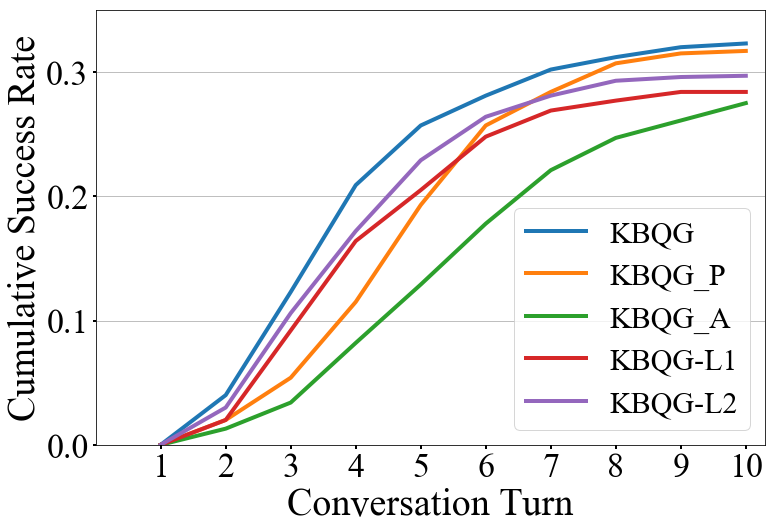}
	}
	\subfigure{}{
		\centering
		\includegraphics[width=4.5cm]{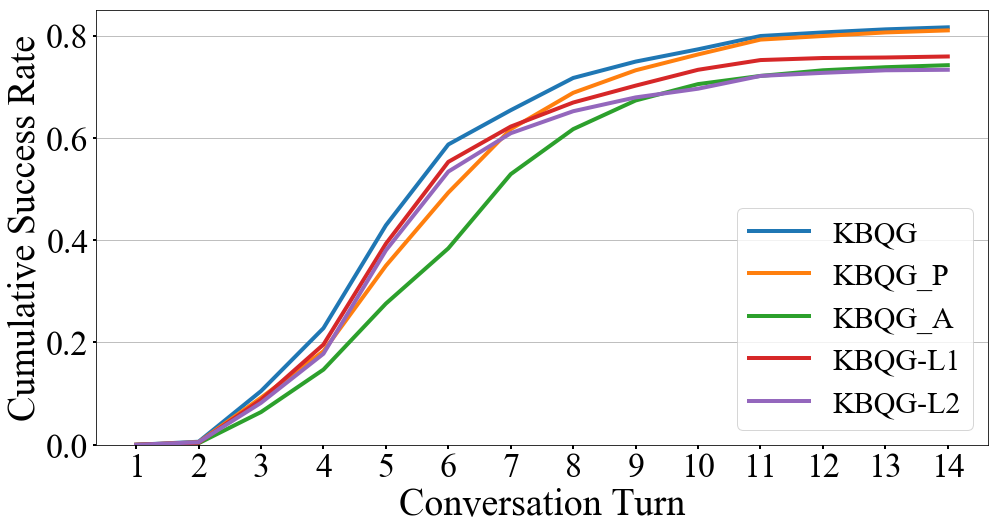}
	}
	\caption{Cumulative success rate of compared methods at different conversation turns on DBbook2014 (left) and MovieLens-1M (right).}
	\label{fig:ablation}
\end{figure}

As can be observed from Figure 5, our proposed KBQG still keeps the best performance over these two variants. The performance drop on each variant proves the effectiveness of our proposed method. KBQG-P shows inferior performance at the early stage of the conversation. This is because KBQG-P needs to train a policy agent to simulate the dialogue distribution, which is easily influenced by the quality of the training corpus, especially when the training corpus is synthesized by a simulated user. Therefore, it cannot distinguish which relation is more important to the user decision, and can only deduce the next dialogue action conditioned on the current dialogue belief state. KBQG-A randomly selects relations to formulate clarifying questions during the conversation, as each relation share the same attention weight for influencing the user decision. This schema may let the system ask clarifying questions about the unimportant relations, which would cripple the model efficiency. 

In addition, we further analyze the influence of joint learning parameter $\lambda$ in Section 4.2 to the overall model performance. By default, we set $\lambda$ to 0.5 on MovieLens-1M and 0.7 on DBbook2014. We alternate the value of $\lambda$ of KBQG, where $\textbf{KBQG-L1}$ uses $\lambda=0.3$ on both datasets, and $\textbf{KBQG-L2}$ uses $\lambda = 0.7$ on MovieLense, and $\lambda=0.5$ on DBbook2014. As we can tell from Figure \ref{fig:ablation}, KBQG-L1 and KBQG-L2 perform worse than our default setting. 
KBQG-L1 performs worse on DBbook2014 but obtains better performance on MovieLens-1M compared with KBQG-L2. The main reason is that MovieLens-1M contains more interaction records for the recommender to capture implicit user preferences. Putting a higher weight on the relation modelling could enhance the entity embeddings to improve the recommendation performance. Whereas DBbook2014 exhibits fewer historical records for modelling the user preference, it is more suitable to lay more emphasis on the recommendation loss, which would benefit the overall recommendation performance. Both the KG relation modelling and recommendation parts simultaneously take effects on the eventual recommendation quality, and $\lambda$ should be adjusted according to the property of each dataset for optimal recommendation performance.

\subsection{Qualitative Analysis (RQ3)}
We further conduct qualitative analysis to show how KBQG is able to improve its understanding of the user preferences and generate appropriate questions for explainable recommendations. We randomly select one real conversation record from KBQG and SCPR on MovieLens based on the same test instance. SCPR is a representative of graph-based approaches that designs its action space based on all attribute values. The conversation is initiated by the user (ID 4607) requesting for a movie recommendation. 

Figure \ref{fig:dialoguedemo} demonstrates the conversations between the user and KBQG/SCPR. It also visualizes the user's attention weights on different relations learned by KBQG. Clearly, KBQG asks clarifying questions following the rank of attention weights on different relations to collect user preference information. From the conversations, we can find KG-based method that relies on the user's feedback to find the recommendation path is substantially less efficient when the dataset has a large attribute pool. Hence, SCPR requires a longer conversation with the user to find a suitable recommendation. In contrast, our proposed KBQG is designed to generate the conversation in a totally different fashion. It substantially uses the recommender system to rank the attention weight of different relations, formulating the most appropriate clarifying questions for obtaining a user's preferences. It is well-suited to the large KG dataset. With the help of attention scores learned on different relations, KBQG is able to filter unnecessary clarifying questions to improve the conversation efficiency. 

\section{Conclusion}
In this paper, we redefine the conversational recommender system and propose a novel KG-based conversational recommender system, KBQG, where the recommender system and the dialogue system closely cooperate with each other so as to efficiently and accurately generate recommendations in a short conversation. Specifically, the preference mining module in KBQG mainly extracts rich auxiliary information from the KG to explicitly explore users preferences from historical records. Conditioned on the explored preference, KBQG can effectively tailor the clarifying questions by priortizing attribute types that are important to the user. After multiple conversation turns, personalized recommendations will be given when the user has sufficiently clarified her/his real-time interests. With extensive experiments on two real-world datasets, KBQG presents superior performance compared with state-of-the-art policy-based and knowledge-based CRSs.

\begin{acks}
This work was supported by ARC Discovery Project (Grant No. DP190101985) and ARC Training Centre for Information Resilience (Grant No. IC200100022). 

\end{acks}

\bibliographystyle{ACM-Reference-Format}
\newpage
\bibliography{sample-bibliography}

\end{document}